\documentclass[letterpaper, 10 pt, conference]{ieeeconf}  

\IEEEoverridecommandlockouts                              

\usepackage{graphicx} 
\usepackage{amsmath,amssymb}
\usepackage{cite}
\usepackage{subfigure}
\usepackage{algorithm}
\usepackage[noend]{algpseudocode}

\title{\LARGE \bf
Making Curiosity Explicit in Vision-based RL
}

\author{Elie Aljalbout*$^{1}$, Maximilian Ulmer*$^{1}$  and Rudolph Triebel$^{1,2}$
\thanks{$^{*}$Shared first authorship \newline
        $^{1}$Technical University of Munich (TUM), 80797 Munich, Germany 
        {\tt\small \{elie.aljalbout@tum.de, max.ulmer@tum.de\}}, 
        $^{2}$German Aerospace Center (DLR), 82234 We\ss ling, Germany
        {\tt\small \{rudolph.triebel@dlr.de\}},
        }
        }

\begin{document}

\maketitle
\thispagestyle{empty}
\pagestyle{empty}

\begin{abstract}
Vision-based reinforcement learning (RL) is a promising technique to solve control tasks involving images as the main observation. State-of-the-art RL algorithms still struggle in terms of sample efficiency, especially when using image observations. This has led to an increased attention on integrating state representation learning (SRL) techniques into the RL pipeline. Work in this field demonstrates a substantial improvement in sample efficiency among other benefits. However, to take full advantage of this paradigm, the quality of samples used for training plays a crucial role. More importantly, the diversity of these samples could affect the sample efficiency of vision-based RL, but also its generalization capability. In this work, we present an approach to improve the sample diversity. Our method enhances the exploration capability of the RL algorithms by taking advantage of the SRL setup. Our experiments show that the presented approach outperforms the baseline for all tested environments. These results are most apparent for environments where the baseline method struggles. Even in simple environments, our method stabilizes the training, reduces the reward variance and boosts sample efficiency.
\end{abstract}

\section{INTRODUCTION}
In order to solve complex tasks in unstructured environments, agents should be capable of learning new skills based on their understanding of their surroundings. Vision-based reinforcement learning is a promising technique to enable such an ability. These methods learn mappings from pixels to actions and may require millions of samples to converge, especially for physical control tasks~\cite{barth-maron2018distributional}. This sample inefficiency could be attributed to the complexity of the dynamics encountered in such environments, but also to the difficulty of processing raw image information. One popular paradigm to approach the latter problem is to integrate state representation learning objectives in the RL process~\cite{ballard1987modular, yarats2019improving}. In contrast to end-to-end methods, approaches that leverage SRL explicitly encourage the policy to learn a state representation mapping based on observations. The additional objective improves sample efficiency as it provides extra signal for training. However, during RL, the agent performs several trials to achieve a certain behavior. This trial and error process, together with the exploitative nature of RL algorithms could result in very similar samples being collected in the replay buffer. This lack of diversity could harm the generalization capability of the learned encoders and also hinder the improvement in sample efficiency that could be achieved with SRL. Hence, data diversity could be very beneficial for vision-based RL and exploration strategies tailored for diverse observations could boost sample efficiency even further. In this work, we aim at improving sample diversity of vision-based RL. We present an approach for exploration that makes the agent specifically curious about the state representation. Our approach takes advantage of the off-policy property of most state-of-the-art RL algorithms and trains a separate exploration policy based on the SRL error. Our experiments show that the proposed method improves the performance of downstream tasks, especially for environments where recent approaches struggle. It also stabilizes the training and reduces the reward variance for all environments.

\section{RELATED WORK}
\label{sec:rel_work}
\textbf{Integrating SRL.} One popular approach for SRL is the autoencoder (AE)~\cite{ballard1987modular}. One of the earliest work to integrate AEs in batch-RL can be found in~\cite{lange2010deep}. Later work explored the use of variational AEs~\cite{higgins2017darla} as well as regularized autoencoders (RAE)~\cite{yarats2019improving}. More recent approaches take advantage of contrastive learning to boost the sample efficiency of vision-based RL. Namely, \cite{laskin2020curl} uses data augmentations as positive samples and all other samples in the batch, as well as their augmentations as the negative ones. Similarly, the work in~\cite{stooke2020decoupling} uses contrastive learning to associate pairs of observations separated by a short time difference, hence uses (near) future observations as positive queries and all other samples in the batch as negative ones. Both AE-based methods and contrastive learning ones focus on compression of observation as the main goal for SRL. Other approaches use dynamic models~\cite{van2016stable}, state-space models~\cite{ha2018world} or domain priors~\cite{jonschkowski2015learning}. In this work, we focus on AE-based approaches for SRL. Although contrastive learning has shown superior results to these approaches~\cite{laskin2020curl, stooke2020decoupling}, AE-based methods have many advantages. They are simple to implement, allow for integrating self-supervised objectives such as jigsaw puzzle~\cite{noroozi2016unsupervised}, enable multi-modal and multi-view fusion~\cite{lee2019making, akinola2020learning}, as well as task specific objectives such as contact prediction~\cite{lee2019making}. More importantly, AE-based approaches lead to a more explainable state representation, especially when using generative AEs.
\begin{figure*}[htb!]
    \centering
    \includegraphics[width=\textwidth]{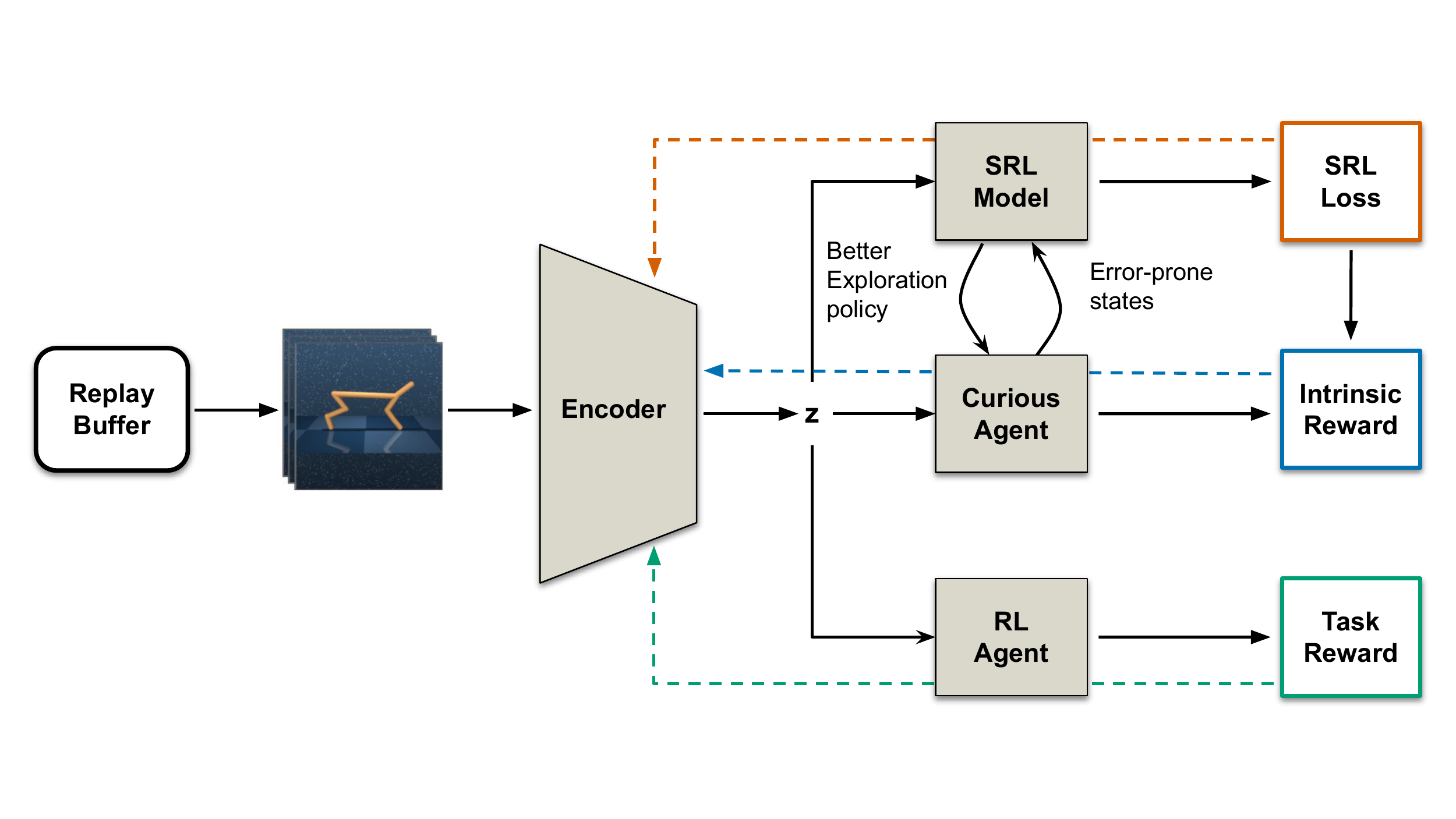}
    \caption{System Overview: our architecture is similar to the classical ones used for simulatenous state representation (SRL) and reinforcement learning (RL). In addition to the classical components, our method introduces a novel curious agent/policy, which is trained based on the SRL loss as an intrinsic reward. This creates an interplay between the SRL and the exhibited curious exploration behaviour. The SRL guides the updates of the curiosity component, while the latter takes actions which lead to problematic and error-prone states. This in turns increase the diversity of observations.}
    \label{fig:Overview}
\end{figure*}
\\
\\
\textbf{Curiosity in RL.}
Classical deep RL algorithms work well in environments with rewards that are easy to encounter, but tend to fail once high-reward areas are harder to reach~\cite{burda2018exploration}.  This clearly motivates the use of exploration techniques as a mean to achieve this goal. Popular paradigms for exploration include counts and pseudo-counts~\cite{DBLP:conf/nips/BellemareSOSSM16,ostrovski2017count}, learning distributions over value functions or policies~\cite{osband2016deep}, and information gain based methods~\cite{schmidhuber1991curious, HouthooftCCDSTA16,pathak2017curiosity}. Our work is mostly related to this last category of methods. While most of these approaches depend on the prediction error of a  dynamics model, we leverage the SRL error for intrinsic motivation. This enables a seamless integration of exploration in vision-based RL without any need for training dynamics models in the process. Furthermore, this creates an interplay between the SRL and the exploration which are both crucial aspects of successful and sample efficient vision-based RL. Most closely related to our method is the work in~\cite{seo2021state}. This work attempts to maximize state entropy using random convolutional encoders. The method uses a k-nearest neighbor entropy estimator in the representation space and uses this estimation as an additional intrinsic reward bonus for RL. Similar to our work, their approach doesn't require any dynamics models for training. However, using a k-nearest neighbor entropy estimator could be either compute-expensive if all observations need to be embedded at each step, or memory-expensive when those embeddings are saved in the replay buffer. Furthermore, a random encoder doesn't guarantee any notion of meaningful similarity between observations. In fact, in certain degenerate cases, the similarity in the representation space of a random encoder could be a measure of dissimilarity of the states. 

\begin{figure*}[htb!]
    \centering
    \subfigure[Finger Spin]{\includegraphics[width=0.32\textwidth]{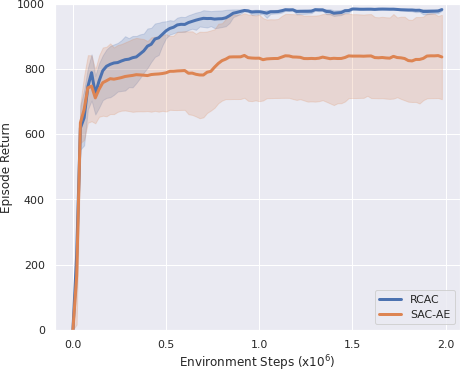}}
    \subfigure[Cartpole Swingup]{\includegraphics[width=0.32\textwidth]{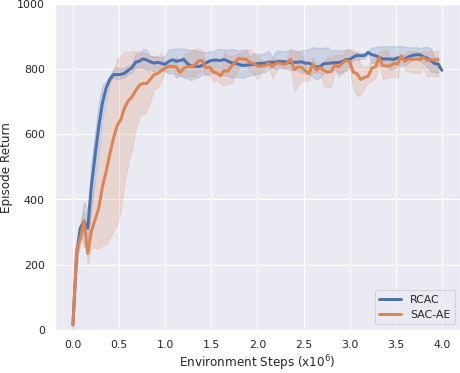}}
    \subfigure[Ball in Cup Catch]{\includegraphics[width=0.32\textwidth]{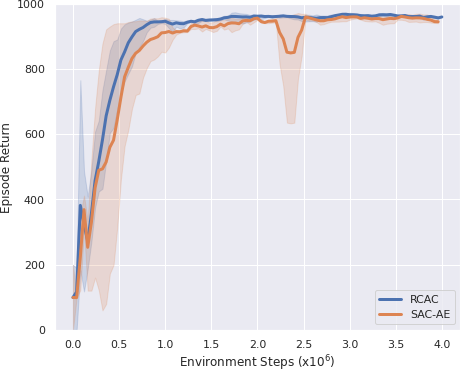}}
    \subfigure[Reacher Easy]{\includegraphics[width=0.32\textwidth]{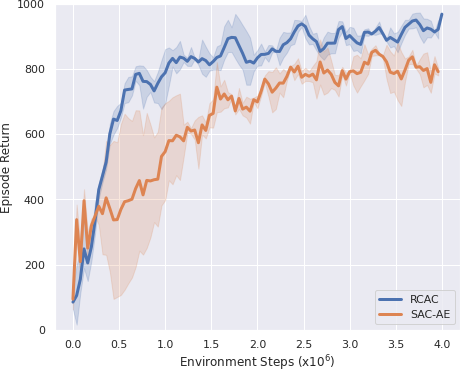}}
    \subfigure[Reacher Hard]{\includegraphics[width=0.32\textwidth]{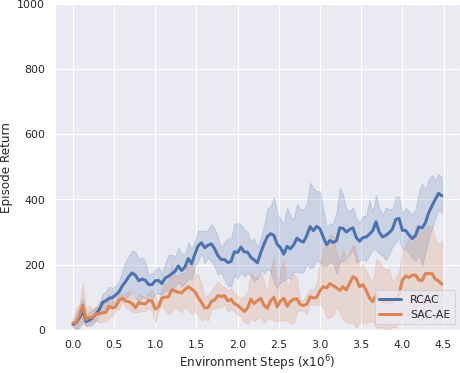}}
    \subfigure[Finger Turn]{\includegraphics[width=0.32\textwidth]{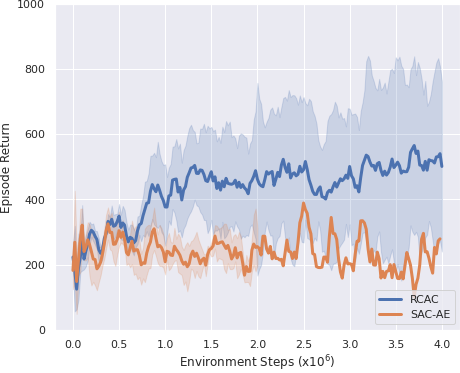}}
    \caption{Training curves on 6 continuous control tasks from the Deepmind Control Suite~\cite{tassa2018deepmind}. In all environments, our method exceeds the performance of the baseline. For easier tasks, the curious exploration either stabilizes the training or improves the maximum achieved reward. For the more difficult tasks (e) and (f), the additional curiosity objective allows to improve the average reward, where the baseline fails to reach high-reward areas.}
    \label{fig:results}
\end{figure*}
\section{APPROACH}

Our main goal is to improve sample diversity of vision-based RL as to boost both the SRL part and the policy learning. The simplest way to approach this problem is to use previous techniques for exploration as the ones presented in section~\ref{sec:rel_work}. However, we argue that having an exploration method that is tailor-made for SRL and takes advantage of its properties can have a positive effect on the overall framework. This paper presents such a method. We propose to use the SRL error as an intrinsic reward $r_{cure}$ for curiosity-based exploration. This reward could either be added to the task reward $r_{task}$ to train the main policy $\pi_{task}$, or used separately for training a separate curious policy $\pi_{cure}$. Previous methods mostly use the earlier approach to integrate the intrinsic reward (based on a dynamics model)~\cite{pathak2017curiosity}. However, our early experiments indicate that a separate curious policy leads to substantially higher reward areas, while the single policy approach could deteriorate the results in comparison to the baseline. Furthermore, adding the rewards together usually introduces extra hyperparameters to weight the different terms (e.g. in~\cite{pairet2019learning}, 3 extra hyperparameters are needed). Hence, in this work, we use two separate policies and train each with either the task reward or the intrinsic one. The update of both policies affect the encoder parameters $\phi$. The SRL model parameters $\theta$  are only affected by the SRL update. Note that the SRL model could refer to different things depending on the SRL approach used. For instance when using an AE-based method, it would correspond to a decoder. The overall architecture is illustrated in figure~\ref{fig:Overview}. At every step, we either sample actions from the main policy or the curious one. The choice of which policy to use at every step is based on a hyperparameter $p_{c}$ which specifies the percentage of times exploration actions should be sampled.  Intuitively, the curious policy is trained to reach states which have high SRL error. By occasionally sampling actions from this policy, the replay buffer ends up containing more problematic and diverse samples which helps to learn a better representation and to avoid overfitting. Subsequently, this could improve the performance of the downstream task. This interaction between the curious policy and the SRL model/loss results in an interplay similar to the one observed in generative adversarial networks~\cite{goodfellow2014generative}, as both modules are mutually beneficial to each other, and are trained in an adversarial setting. This interplay is illustrated in figure~\ref{fig:Overview}. It is important to note that having a separate policy is only possible when using off-policy RL algorithms. In this work we use an RAE for SRL and soft actor critic (SAC)~\cite{haarnoja2018soft} for RL. This combination is explored in~\cite{yarats2019improving}. The RAE is trained with the following objective:
\begin{equation}
\label{eq:rae}
    J(RAE) = \mathbb{E}_{o_t\sim D}[\log p_\theta(o_t|z_t) + \lambda_z ||z_t||^2 + \lambda_\theta||\theta||^2 ] 
\end{equation}

where $z_t=g_\phi(o_t)$ is the learned state representation, $o_t$ is the image observation, $g_\phi$ is the encoder, and $\lambda_z$, $\lambda_\theta$ are hyperparameters which respectively specify the influence of the $L_2$ penalty on $z$ and the weights decay for the decoder parameters.

During training, at every step, the intrinsic reward is computed as $r_{cure}=J(RAE)$. We refer to the overall algorithm as RCAC, which stands for: Representation-Curious Actor-Critic. RCAC is summarized in algorithm~\ref{pseudo}. 

\begin{algorithm}
\caption{\textbf{RCAC}: \textbf{R}epresentation-\textbf{C}urious \textbf{A}ctor-\textbf{C}ritic}\label{pseudo}
\begin{algorithmic}
\For{ each timestep $t=1...T$}
    \State $\epsilon \sim U(0,1)$
    \If{$\epsilon < p_c$}
        \State $a_t \sim \pi_{cure}(.|o_t)$
    \Else 
        \State $a_t \sim \pi_{task}(.|o_t)$
    \EndIf
    \State $o_t'\sim p(.|o_t,a_t)$
    \State $D \gets D\cup (o_t,a_t,r_{task}(o_t,a_t),o_t')$
    \State $B \gets SampleBatch(D)$
    \State $r_{cure} \gets UpdateRAE(B)$ \Comment{equation~(\ref{eq:rae})}
    \State $UpdateTaskAC(B)$ \Comment{using $r_{task}$}
    \State $UpdateCuriousAC(B, r_{cure})$ \Comment{using  $r_{cure}$}
\EndFor
\end{algorithmic}
\end{algorithm}

\section{EXPERIMENTS} 
\textbf{Setup.} We experimentally evaluate our method on six continuous control tasks from the Deepmind Control Suite~\cite{tassa2018deepmind}. The chosen tasks aim to cover a wide range of common RL challenges, such as contact dynamics and sparse rewards. The tasks we use are \texttt{reacher\_easy}, \texttt{cartpole\_swingup}, \texttt{ball\_in\_cup\_catch}, \texttt{finger\_spin}, \texttt{finger\_turn} and \texttt{reacher\_hard}. To evaluate the benefits of curious exploration via the SRL loss, we compare our method to SAC-AE~\cite{yarats2019improving} as a baseline, on which our implementation is based. For simplicity, we use the same hyperparameters for all experiments except of the action repeat value which changes per task, according to~\cite{hafner2019learning}. The actor and critic networks for the RL agent and the curious agent are trained using the Adam optimizer~\cite{kingma2014adam}, using default parameters. We store trajectories of experiences in a standard replay buffer. For the SAC implementation, we follow the training procedure of SAC detailed in~\cite{yarats2019improving}. For the sake of reproduciblity, we provide an overview of the hyper parameters in Table~\ref{tab:hyper}. 

\begin{table}
\normalsize
\begin{center}
\begin{tabular}{ l|c } 
 \hline
 Parameter & Setting \\ 
 \hline
 Batch size & 128 \\ 
 Replay buffer capacity & 80000 \\ 
 Discount $\gamma$ & 0.99 \\
 Hidden dimension & 1024 \\
 Curios exploration probability $p_c$ & 0.2 \\
 Observation size & $84\times84\times3$ \\
 Frames stacked & 3 \\
 Critic learning rate & $10^{-3}$ \\
 Critic target update frequency & 2 \\
 Critic soft target update rate $\tau$ & 0.01 \\
 Actor learning rate & $10^{-3}$ \\
 Actor update frequency & 2 \\
 Actor log std bounds & [-10, 2] \\
 Autoencoder learning rate & $10^{-3}$ \\
 Decoder update frequency & 1 \\
 Temperature learning rate & $10^{-4}$ \\
 
 Init temperature & 0.1 \\
 
 \hline
\end{tabular}
\caption{The hyperparameters used in our experiments.}
\label{tab:hyper}
\end{center}
\end{table}

For the encoder and decoder, we employ the architecture from~\cite{yarats2019improving}. Both consist of four convolutional layers with $3 \times 3$ kernels and $32$ channels and use ReLU activations, except of the final deconvolution layer. Both networks use a stride of 1 for each layer except of the first of the encoder and the last of the decoder, which use stride 2.

To evaluate the performance of our model, we train multiple different seeds for each task. At the beginning of each seed, we pretrain the models with 1000 samples which we collect by rolling-out random actions. Afterwards, we evaluate the model every 10000 environment steps over 10 episodes and report the average reward. The results are shown in Figure~\ref{fig:results}. The total number of episodes depends on the complexity of the task.
\\
\\
\textbf{Results.}
Figure~\ref{fig:results} shows the task reward achieved by RCAC in comparison to the baseline. In all environments, our method exceeds the performance of the baseline implementation. Specifically, for tasks where the baseline doesn't show any signs of improvement, such as \texttt{reacher\textunderscore hard} and \texttt{finger\textunderscore turn}, RCAC leads to exploring high-reward areas, as can be seen when looking at the maximum rewards achieved in those environments. For simpler tasks such as \texttt{reacher\textunderscore easy} and \texttt{finger\textunderscore spin}, RCAC approaches the maximum environment rewards, while the baseline converges to $80\%$. In addition to this boost in achieved reward, RCAC stabilizes the training and reduces the reward variance significantly. This feature is usually not given enough attention in RL research. However, in real-world scenarios, when deploying RL agents for real tasks, there could be cases where only one training run is possible. An algorithm with lower reward variance could guarantee a sufficiently good policy, while it's hard to say the same when this condition fails. This effect can also be seen for \texttt{cartpole\textunderscore swingup} and \texttt{ball\textunderscore in\textunderscore cup\textunderscore catch}. RCAC seems to have a minor effect on the maximum reached reward for these last 2 environments. This could be attributed to the fact that for these environments, the baseline already approaches the performance achieved by SAC trained with the true states~\cite{yarats2019improving}. The additional curious exploration objective, accelerates the convergences of all evaluation tasks, thus improving the sample efficiency which is one key limitation of state-of-the-art model-free algorithms. In general, it could be seen from our experiments that RCAC becomes more effective when the task complexity increase. This will be especially important, moving towards more
complex tasks. We plan to rigorously test this hypothesis in future experiments. 

\section{CONCLUSION}
In this work, we introduce a curiosity-based exploration technique which can be easily used together with state representation learning methods used in RL. This method exploits the SRL error to incentivize visiting more diverse and problematic states. Our experiments show that our curious exploration method improves the performance of SAC-AE trained from pixels on complex continuous control tasks. When comparing SAC-AE to curious SAC-AE on six tasks of the deepmind control suite, we show that the added curiosity improves the performance in terms of convergence, stability and the achieved total reward. In future work, we plan to experiment with the transfer learning capability of our architecture, scale it up to even harder tasks, test the effect of the exploration hyperparameter, and test our approach with different SRL methods.






\bibliographystyle{plain}
\bibliography{bibliography}

\begin{thebibliography}{10}

\bibitem{akinola2020learning}
Iretiayo Akinola, Jacob Varley, and Dmitry Kalashnikov.
\newblock Learning precise 3d manipulation from multiple uncalibrated cameras.
\newblock In {\em 2020 IEEE International Conference on Robotics and Automation
  (ICRA)}, pages 4616--4622. IEEE, 2020.

\bibitem{ballard1987modular}
Dana~H Ballard.
\newblock Modular learning in neural networks.
\newblock In {\em AAAI}, pages 279--284, 1987.

\bibitem{barth-maron2018distributional}
Gabriel Barth-Maron, Matthew~W. Hoffman, David Budden, Will Dabney, Dan Horgan,
  Dhruva TB, Alistair Muldal, Nicolas Heess, and Timothy Lillicrap.
\newblock Distributional policy gradients.
\newblock In {\em International Conference on Learning Representations}, 2018.

\bibitem{DBLP:conf/nips/BellemareSOSSM16}
Marc~G. Bellemare, Sriram Srinivasan, Georg Ostrovski, Tom Schaul, David
  Saxton, and Rémi Munos.
\newblock Unifying count-based exploration and intrinsic motivation.
\newblock In {\em NIPS}, pages 1471--1479, 2016.

\bibitem{burda2018exploration}
Yuri Burda, Harrison Edwards, Amos Storkey, and Oleg Klimov.
\newblock Exploration by random network distillation.
\newblock In {\em International Conference on Learning Representations}, 2019.

\bibitem{goodfellow2014generative}
Ian~J Goodfellow, Jean Pouget-Abadie, Mehdi Mirza, Bing Xu, David Warde-Farley,
  Sherjil Ozair, Aaron Courville, and Yoshua Bengio.
\newblock Generative adversarial nets.
\newblock In {\em Proceedings of the 27th International Conference on Neural
  Information Processing Systems-Volume 2}, pages 2672--2680, 2014.

\bibitem{ha2018world}
David Ha and J{\"u}rgen Schmidhuber.
\newblock World models.
\newblock {\em arXiv preprint arXiv:1803.10122}, 2018.

\bibitem{haarnoja2018soft}
Tuomas Haarnoja, Aurick Zhou, Pieter Abbeel, and Sergey Levine.
\newblock Soft actor-critic: Off-policy maximum entropy deep reinforcement
  learning with a stochastic actor.
\newblock In {\em International Conference on Machine Learning}, pages
  1861--1870. PMLR, 2018.

\bibitem{hafner2019learning}
Danijar Hafner, Timothy Lillicrap, Ian Fischer, Ruben Villegas, David Ha,
  Honglak Lee, and James Davidson.
\newblock Learning latent dynamics for planning from pixels.
\newblock In {\em International Conference on Machine Learning}, pages
  2555--2565. PMLR, 2019.

\bibitem{higgins2017darla}
Irina Higgins, Arka Pal, Andrei Rusu, Loic Matthey, Christopher Burgess,
  Alexander Pritzel, Matthew Botvinick, Charles Blundell, and Alexander
  Lerchner.
\newblock Darla: Improving zero-shot transfer in reinforcement learning.
\newblock In {\em International Conference on Machine Learning}, pages
  1480--1490. PMLR, 2017.

\bibitem{HouthooftCCDSTA16}
Rein Houthooft, Xi~Chen, Yan Duan, John Schulman, Filip~De Turck, and Pieter
  Abbeel.
\newblock Vime: Variational information maximizing exploration.
\newblock In {\em NIPS}, pages 1109--1117, 2016.

\bibitem{jonschkowski2015learning}
Rico Jonschkowski and Oliver Brock.
\newblock Learning state representations with robotic priors.
\newblock {\em Autonomous Robots}, 39(3):407--428, 2015.

\bibitem{kingma2014adam}
Diederik~P Kingma and Jimmy Ba.
\newblock Adam: A method for stochastic optimization.
\newblock {\em arXiv preprint arXiv:1412.6980}, 2014.

\bibitem{lange2010deep}
Sascha Lange and Martin Riedmiller.
\newblock Deep auto-encoder neural networks in reinforcement learning.
\newblock In {\em The 2010 International Joint Conference on Neural Networks
  (IJCNN)}, pages 1--8. IEEE, 2010.

\bibitem{laskin2020curl}
Michael Laskin, Aravind Srinivas, and Pieter Abbeel.
\newblock Curl: Contrastive unsupervised representations for reinforcement
  learning.
\newblock In {\em International Conference on Machine Learning}, pages
  5639--5650. PMLR, 2020.

\bibitem{lee2019making}
Michelle~A Lee, Yuke Zhu, Krishnan Srinivasan, Parth Shah, Silvio Savarese,
  Li~Fei-Fei, Animesh Garg, and Jeannette Bohg.
\newblock Making sense of vision and touch: Self-supervised learning of
  multimodal representations for contact-rich tasks.
\newblock In {\em 2019 International Conference on Robotics and Automation
  (ICRA)}, pages 8943--8950. IEEE, 2019.

\bibitem{noroozi2016unsupervised}
Mehdi Noroozi and Paolo Favaro.
\newblock Unsupervised learning of visual representations by solving jigsaw
  puzzles.
\newblock In {\em European conference on computer vision}, pages 69--84.
  Springer, 2016.

\bibitem{osband2016deep}
Ian Osband, Charles Blundell, Alexander Pritzel, and Benjamin Van~Roy.
\newblock Deep exploration via bootstrapped dqn.
\newblock {\em arXiv preprint arXiv:1602.04621}, 2016.

\bibitem{ostrovski2017count}
Georg Ostrovski, Marc~G Bellemare, A{\"a}ron Oord, and R{\'e}mi Munos.
\newblock Count-based exploration with neural density models.
\newblock In {\em International conference on machine learning}, pages
  2721--2730. PMLR, 2017.

\bibitem{pairet2019learning}
{\`E}ric Pairet, Paola Ard{\'o}n, Frank Broz, Michael Mistry, and Yvan
  Petillot.
\newblock Learning and generalisation of primitives skills towards robust
  dual-arm manipulation.
\newblock {\em arXiv preprint arXiv:1904.01568}, 2019.

\bibitem{pathak2017curiosity}
Deepak Pathak, Pulkit Agrawal, Alexei~A Efros, and Trevor Darrell.
\newblock Curiosity-driven exploration by self-supervised prediction.
\newblock In {\em International Conference on Machine Learning}, pages
  2778--2787. PMLR, 2017.

\bibitem{schmidhuber1991curious}
Juergen Schmidhuber.
\newblock Curious model-building control systems.
\newblock In {\em Proc. international joint conference on neural networks},
  pages 1458--1463, 1991.

\bibitem{seo2021state}
Younggyo Seo, Lili Chen, Jinwoo Shin, Honglak Lee, Pieter Abbeel, and Kimin
  Lee.
\newblock State entropy maximization with random encoders for efficient
  exploration.
\newblock {\em arXiv preprint arXiv:2102.09430}, 2021.

\bibitem{stooke2020decoupling}
Adam Stooke, Kimin Lee, Pieter Abbeel, and Michael Laskin.
\newblock Decoupling representation learning from reinforcement learning.
\newblock {\em arXiv preprint arXiv:2009.08319}, 2020.

\bibitem{tassa2018deepmind}
Yuval Tassa, Yotam Doron, Alistair Muldal, Tom Erez, Yazhe Li, Diego de~Las
  Casas, David Budden, Abbas Abdolmaleki, Josh Merel, Andrew Lefrancq, et~al.
\newblock Deepmind control suite.
\newblock {\em arXiv preprint arXiv:1801.00690}, 2018.

\bibitem{van2016stable}
Herke Van~Hoof, Nutan Chen, Maximilian Karl, Patrick van~der Smagt, and Jan
  Peters.
\newblock Stable reinforcement learning with autoencoders for tactile and
  visual data.
\newblock In {\em 2016 IEEE/RSJ international conference on intelligent robots
  and systems (IROS)}, pages 3928--3934. IEEE, 2016.

\bibitem{yarats2019improving}
Denis Yarats, Amy Zhang, Ilya Kostrikov, Brandon Amos, Joelle Pineau, and Rob
  Fergus.
\newblock Improving sample efficiency in model-free reinforcement learning from
  images.
\newblock {\em arXiv preprint arXiv:1910.01741}, 2019.

\end{thebibliography}

\end{document}